\documentclass{article}

\usepackage[final, nonatbib]{neurips_2021}

\usepackage[utf8]{inputenc}
\usepackage[T1]{fontenc}
\usepackage{hyperref}
\usepackage{url}
\usepackage{booktabs}
\usepackage{amsfonts}
\usepackage{nicefrac}
\usepackage{microtype}

\usepackage{amsmath}
\usepackage{amssymb}
\usepackage{siunitx}
\usepackage{todonotes}
\usepackage{multirow}

\title{Physics-informed inference of aerial animal movements from weather radar data}

\author{%
  Fiona~Lippert\thanks{Correspondence: \texttt{f.lippert@uva.nl}} \\
  AI4Science Lab, AMLab, IBED \\
  University of Amsterdam\\
  \And
  Bart Kranstauber \\
  IBED \\
  University of Amsterdam \\
  \AND
  E. Emiel van Loon \\ 
  IBED \\
  University of Amsterdam \\
  \And
  Patrick Forr\'e \\
  AI4Science Lab, AMLab \\
  University of Amsterdam \\
}

\begin{document}

\maketitle

\begin{abstract}
  Studying animal movements is essential for effective wildlife conservation and conflict mitigation. For aerial movements, operational weather radars have become an indispensable data source in this respect. However, partial measurements, incomplete spatial coverage, and poor understanding of animal behaviours make it difficult to reconstruct complete spatio-temporal movement patterns from available radar data.
  We tackle this inverse problem by learning a mapping from high-dimensional radar measurements to low-dimensional latent representations using a convolutional encoder.
  Under the assumption that the latent system dynamics are well approximated by a locally linear Gaussian transition model, we perform efficient posterior estimation using the classical Kalman smoother.
  A convolutional decoder maps the inferred latent system states back to the physical space in which the known radar observation model can be applied, enabling fully unsupervised training.
  To encourage physical consistency, we additionally introduce a physics-informed loss term that leverages known mass conservation constraints.
  Our experiments on synthetic radar data show promising results in terms of reconstruction quality and data-efficiency.
\end{abstract}

\section{Introduction}

Doppler weather radars provide high-resolution information about the distribution and movement of objects in the atmosphere.
Although designed to monitor weather, radar beams are reflected not only by precipitation but also by animals passing the airspace around the radar.
This offers invaluable opportunities for ecologists to study mass movements of birds, bats and insects that otherwise remain hidden due to low light conditions and high flight altitudes \cite{Bauer2017}.
However, since the measurement range of radars is limited, the spatial coverage of operational weather radar networks is typically incomplete \cite{Kranstauber2020}.
Moreover, while the amount of energy reflected back to the radar antenna can be translated directly into animal density estimates, movements can be captured only partially by measuring the \emph{radial} velocity, i.e. the component of movement along the direction of the radar beam, based on the Doppler shift \cite{doviak2006doppler}. 
Inferring the complete underlying density and velocity fields from partial observations of weather radar networks remains a challenging inverse problem that requires additional knowledge to constrain the solution space.

Unfortunately, the movement of animals is much harder to understand than that of physical particles: individual behaviours depend on a wide range of environmental and social factors, and can vary substantially within a population.
That means that, unlike meteorologists, ecologists are lacking well-established mathematical equations that can be used as dynamical prior in data assimilation frameworks such as \cite{Sun2001, Aksoy2009}. Instead, strong spatio-temporal smoothness assumptions have been used to reconstruct continental-scale movement patterns \cite{Angell2018, Nussbaumer2021}, leaving more fine-scale local movements unresolved.
Machine learning opens up new opportunities in this respect: deep neural networks can learn physically consistent spatio-temporal dynamics from high-dimensional data sets by integrating (partial) domain knowledge into the training process \cite{Kochkov2021, Gao2021, Wang2021a}.
However, the application to ecological systems, and animal movements in particular, is hindered by the size of available data sets (particularly in Europe \cite{ShamounBaranes2021}), as well as the lack of ground-truth data.

Inspired by recent work on combining deep learning with exact inference in linear Gaussian state space models \cite{Becker2019a, Kurle2021, Ruhe2021, Revach2021}, we propose a data-efficient approach to reconstruct high-dimensional time-varying animal density and velocity fields from partial and noisy radar data. 
We jointly model densities and velocities in a learned feature space in which the dynamics are well approximated by a locally linear Gaussian transition model. In this space, inference can be performed with the classical Kalman filter or smoother \cite{Kalman1960, Rauch1965}.
The mapping between physical space and feature space is learned together with the latent dynamics in an unsupervised way, minimizing both the reconstruction loss and a physics-informed loss that exploits known mass conservation constraints based on the continuity equation. 
This allows us to reconstruct high-resolution movement patterns while having (i) incomplete knowledge about the underlying process, (ii) no access to ground-truth fields, and (iii) only a limited number of training samples to learn a surrogate for the complex spatio-temporal dynamics.

\section{Background}\label{sec:data}

In this section, we briefly describe relevant prior knowledge about mass movements of animals and discuss how weather radars can be used to quantify such processes.

\paragraph{Modelling mass movements}
The coordinated movement of large number of animals, e.g. during migration, can be described in terms of individuals moving along a shared velocity field $\mathbf{v}$ that changes over time depending on weather conditions, landscape features, food availability, social interactions, etc \cite{Nussbaumer2021}. The resulting spatio-temporal distribution of animals is restricted by the continuity equation
\begin{equation}\label{eq:transport}
    \frac{\partial \rho}{\partial t} = - \nabla\cdot (\mathbf{v}\rho),
\end{equation}
where $\rho$ denotes animal density. The spatio-temporal dynamics of the velocity field $\mathbf{v}$ and the underlying drivers, however, remain largely unknown and need to be estimated from data.

\paragraph{Radar basics}\label{sec:radar_basics}
For birds, bats and insects, continent-wide networks of Doppler weather radars have become an invaluable tool to quantify mass movements over large spatial and temporal extents \cite{Bauer2017}.
Every 5-15 min, Doppler weather radars perform multiple \SI{360}{\degree} sweeps at different elevation angles to sample the three dimensional airspace around the antenna.
With every sweep, several data products are obtained. In this work, we are specifically interested in \emph{reflectivity}, from which the animal density $\rho$ can be estimated, and \emph{radial velocity}, describing associated movements towards or away from the antenna derived from the Doppler shift of reflected radio waves. More precisely, the radial velocity $ r(\mathbf{x}) = \mathbf{a}(\mathbf{x})^T\mathbf{v}(\mathbf{x})$ of a moving object at location $\mathbf{x}$ is the projection of the full velocity vector $\mathbf{v}(\mathbf{x})$ onto the unit vector $\mathbf{a}(\mathbf{x})$ pointing in the direction of the radar beam.
In general, the geometry of radar measurements in combination with scattering effects restricts data collection to some maximum distance from the antenna.
Hence, we can define the forward model $\mathcal{H}$ mapping the true fields $\mathbf{v}$ and $\rho$ to measurements taken by radar $n$:
\begin{align}\label{eq:forward_model}
    \mathcal{H}(\mathbf{v}, \rho, \mathbf{A}_n, d_n) = \left[f_{<d_{n}}(\mathbf{A}_n\mathbf{v}), f_{<d_{n}}(\rho)\right] = \left[\Tilde{\mathbf{r}}_n, \Tilde{\boldsymbol{\rho}}_n\right]
\end{align}
where $f_{<d_{n}}(\cdot)$ represents the partial (i.e. masked) observation of a quantity up to distance $d_n$ from the radar, and the rows of matrix $\mathbf{A}_n$ contain the unit vectors $\mathbf{a}_n(\mathbf{x})$ associated with radial velocity measurements taken by radar $n$.

\paragraph{Velocity field reconstruction}
Unless measurements of multiple radars are spatially overlapping, the reconstruction of $\mathbf{v}$ from $\{\Tilde{\mathbf{r}}_n\}_{n=1}^N$ presents an under-determined inverse problem.
To constrain the solution space, ecological studies traditionally assume movements to be uniform. The vector field $\mathbf{v}$ then reduces to a single vector which can easily be estimated with \emph{velocity volume profiling} techniques \cite{Waldteufel1979, vanGasteren2008}. In practice, however, animal movements are heterogeneous: local weather conditions, barriers, food availability, etc. lead to high variability in both movement directions and speeds.
Angell et al. \cite{Angell2018} made a first step towards reconstructing spatially detailed velocity fields by simultaneously modelling radial velocity measurements from many spatially distributed radars with a Gaussian process model that encodes smoothness assumptions about the latent velocity field. While this allows ecologists to recover large-scale movements across continents, more fine-scale patterns are neglected.

\section{Method}

\begin{figure}[htb]
    \centering
    \includegraphics[width=\textwidth]{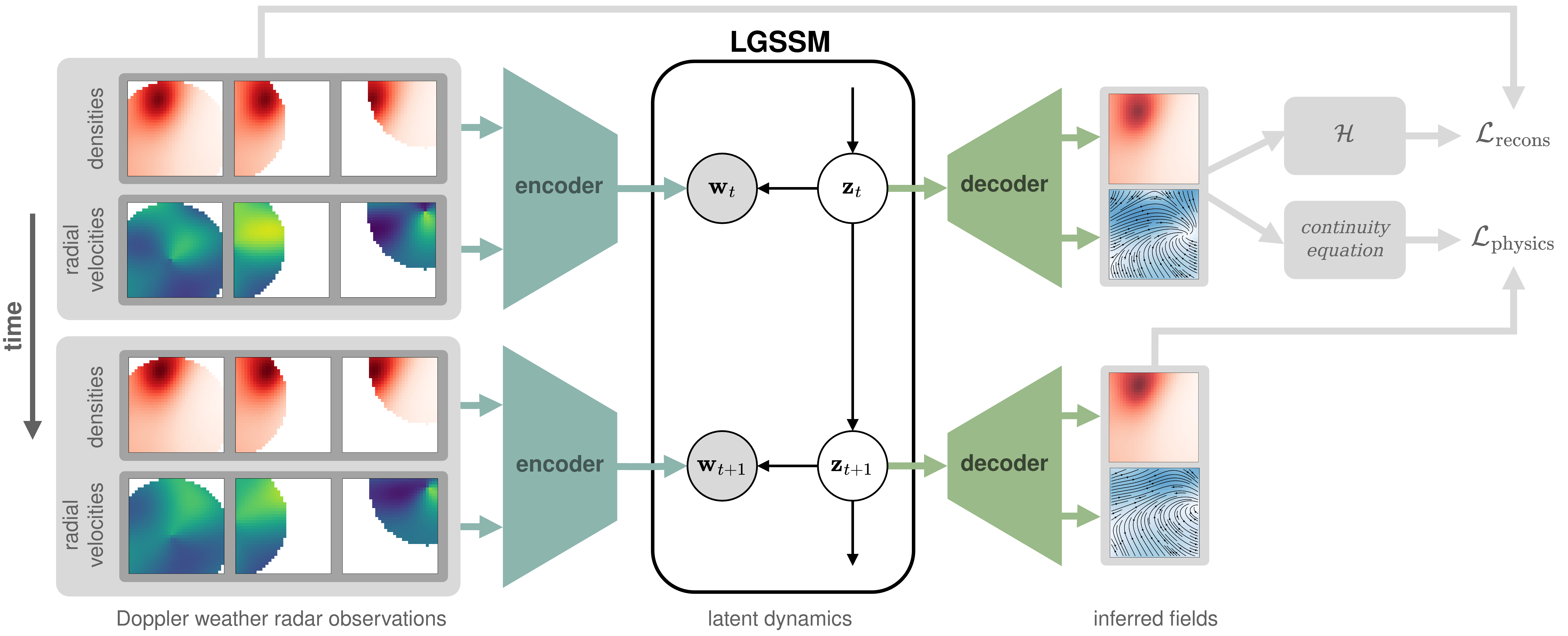}
    \caption{Schematic overview of our method. Partial and noisy observations from Doppler weather radars are mapped to lower-dimensional latent observations $\mathbf{w}_t$, which are used to perform efficient inference in the LGSSM using the Kalman smoother. Finally, posterior estimates for latent states $\mathbf{z}_t$ are mapped back to physical space in which the reconstruction loss and a physics-informed loss are computed.}
    \label{fig:model_overview}
\end{figure}

Consider a time series of radar observations $(\mathbf{o}_1,...,\mathbf{o}_T)$ with $\mathbf{o}_t=\left[\Tilde{\mathbf{r}}_{t}, \Tilde{\mathbf{q}}_{t}\right]$ consisting of radial velocity measurements $\Tilde{\mathbf{r}}_{t}\in\mathbb{R}^{K\times L\times N}$ and log-transformed density measurements $\Tilde{\mathbf{q}}_{t}=\log(\Tilde{\boldsymbol{\rho}}_{t})\in\mathbb{R}^{K\times L\times N}$ on a two-dimensional $K\times L$ grid, taken by $N$ different radars with known projections $\mathbf{A}=\left[\mathbf{A}_1,...,\mathbf{A}_N\right]$ and ranges $\mathbf{d}=\left[d_1,...,d_N\right]$.
Given this partial and noisy data, we aim at reconstructing the true underlying velocity and log-density fields $\{\mathbf{v}_t, \mathbf{q}_t\}_{t=1}^T$.
For a visual summary of our method, see fig. \ref{fig:model_overview}.

\paragraph{Encoding radar measurements}
Inspired by traditional reduced order modelling approaches for high-resolution fluid dynamics \cite{Burkardt2006, Carlberg2011}, we assume that animal movements can be compressed to a much lower dimensional space.
To learn this unknown space from data, we use a convolutional encoder network $f_{\text{enc}}(\mathbf{o}_t, \mathbf{A}) = \mathbf{w}_t$ that maps high-dimensional measurements $\mathbf{o}_t$ together with the associated radar projections $\mathbf{A}$ to a latent measurement $\mathbf{w}_t\in\mathcal{W}\subseteq\mathbb{R}^M$ capturing key spatial features of the measured fields.

\paragraph{Linear Gaussian state space model}
Apart from spatial features, animal movements are characterised by strong temporal dependencies. Moreover, radar measurements can be noisy or contain artefacts due to reflections by other objects in the atmosphere. To account for both temporal dynamics and measurement errors, we model latent measurements $(\mathbf{w}_1,...\mathbf{w}_T)$ with a linear Gaussian state space model (LGSSM)
\begin{equation}\label{eq:LGSSM}
    p(\mathbf{z}, \mathbf{w}) = \mathcal{N}(\mathbf{z}_1\mid \mathbf{\mu}_1, \mathbf{\Sigma}_1) \prod_{t=2}^T \mathcal{N}(\mathbf{z}_t\mid\mathbf{F}_t\mathbf{z}_{t-1}, \mathbf{P}) \prod_{t=1}^T \mathcal{N}(\mathbf{w}_t\mid\mathbf{H}\mathbf{z}_t, \mathbf{R}),
\end{equation}
which links latent measurements to latent representations $\mathbf{z}_t\in\mathcal{Z}\subseteq\mathbb{R}^D$ of the true physical state $(\mathbf{v}_t, \mathbf{q}_t)$.
Here, $\mathbf{H}$ and $\mathbf{P}$ are predefined, $\mathbf{\mu}_1, \mathbf{\Sigma}_1$, and $\mathbf{R}$ are learned parameters, and $\mathbf{F}_t$ is a linear combination of $C$ constant transition matrices $\mathbf{F}^{(k)}$ with coefficients defined by a small neural network that takes the current latent state $\mathbf{z}_t$ as input.

\paragraph{Inference}
Given a sequence of encoded measurements $(\mathbf{w}_1,...,\mathbf{w}_T)$ and the LGSSM from eq.~\ref{eq:LGSSM}, we can perform efficient inference in latent space using the standard Kalman smoother \cite{Rauch1965}.
This results in latent state posterior estimates $\mathbf{z}_1^+,...,\mathbf{z}_T^+$ and corresponding covariances $\mathbf{\Sigma}_1^+,...,\mathbf{\Sigma}_T^+$.
Finally, to recover the physical system states from some latent state $\mathbf{z}_t$, we use a convolutional decoder network to learn the inverse mapping $f_{\text{dec}}(\mathbf{z}_t) = \left[\mathbf{v}_t, \mathbf{q}_t\right]$ from the latent state space $\mathcal{Z}$ back to the physical space $\mathcal{P}\subseteq\mathbb{R}^{K\times L\times 3}$.

\paragraph{Training}
Since in practice it is impossible to obtain data on the true physical states $\{\mathbf{v}_t, \mathbf{q}_t\}_{t=1}^T$, we rely on unsupervised learning. That means, we minimize the reconstruction loss
\begin{equation}
   \mathcal{L}_{\text{recons}} = \frac{1}{T} \sum_{t=1}^T ||\mathbf{o}_t - \mathcal{H}(f_{\text{dec}}(\mathbf{z}^+_t), \mathbf{A}, \mathbf{d})||_2^2,
\end{equation}
where we first map each posterior estimate $\mathbf{z}_t^+$ back to physical space, and then apply the known forward model $\mathcal{H}$ (see eq.~\ref{eq:forward_model}) to obtain the corresponding reconstruction of radar measurements.

\paragraph{Incorporating physical constraints}

To encourage physical consistency, i.e. compliance with conservation laws, we regularize our model based on the continuity equation (see section~\ref{sec:data_generation}). In particular, we map log-densities $\mathbf{q}_t$ back to densities $\rho_t$, and discretize equation~\ref{eq:transport} in time (using forward Euler) and in space (using finite differences):
\begin{align}
    \rho_{t+1}^{(k,l)} - \rho_{t}^{(k,l)} &\approx -\Delta t \cdot \left(\frac{\partial j_{t, x}}{\partial x} + \frac{\partial j_{t, y}}{\partial y}\right)^{(k,l)}\\
    &\approx -\Delta t \cdot \left(\frac{j_{t, x}^{(k+1,l)} - j_{t, x}^{(k-1,l)}}{2\Delta x} + \frac{j_{t, y}^{(k,l+1)} - j_{t, y}^{(k,l-1)}}{2\Delta y} \right),
\end{align}
with $\mathbf{j}_t^{(k,l)}=\left[j_{t,x}^{(k,l)}, j_{t,y}^{(k,l)}\right]^T = \mathbf{v}_t^{(k,l)}\cdot\rho_t^{(k,l)}$ the flux at time $t$ and grid cell $(x_k, y_l)$.
The resulting physical loss term
\begin{equation}
    \mathcal{L}_{\text{physics}} = \frac{1}{T} \sum_{t=1}^T\sum_{k=2}^{K-1}\sum_{l=2}^{L-1} \left\{\rho_{t+1}^{(k,l)} - \rho_{t}^{(k,l)} + \Delta t \left(\frac{\mathbf{j}_{t}^{(k+1,l)} - \mathbf{j}_{t}^{(k-1,l)}}{2\Delta x} + \frac{\mathbf{j}_{t}^{(k,l+1)} - \mathbf{j}_{t}^{(k,l-1)}}{2\Delta y} \right)\right\}
\end{equation}
is combined with the reconstruction loss to form the final training objective
\begin{equation}
    \min_{\theta} \ \mathcal{L}_{\text{recons}} + \mathcal{L}_{\text{physics}},
\end{equation}
where $\theta$ includes encoder, decoder, and LGSSM parameters.

\section{Experiments and results}

\subsection{Synthetic data generation}\label{sec:data_generation}
Besides noisy and partial measurements, working with real weather radar data comes with several additional challenges, such as aliasing, clutter effects, irregular sampling, and discriminating biology from weather. Moreover, the lack of ground truth fields prevents a thorough evaluation of model reconstructions. We thus generate a synthetic radar data set to train and evaluate our proposed methodology (see appendix A for details).
In particular, we generate 1000 training sequences and 50 test sequences of time-varying 2D velocity and animal density fields with a resolution of $32\times 32$ and length $T=20$ (each corresponding to ca. 2-6h of radar data). The associated partial and noisy measurements $\{\Tilde{\boldsymbol{\rho}}_n, \Tilde{\mathbf{r}}_n\}_{n=1}^3$ of 3 randomly positioned radars are obtained by applying the respective forward model and adding independent Gaussian noise with $\sigma=0.001$.

\subsection{Architecture and hyperparameters}
\paragraph{Convolutional encoder and decoder}
In all our experiments, we use a convolutional encoder network consisting of 3 layers with $3\times 3$ filters, $1\times 1$ stride, ReLU non-linearities, and $2\times 2$ max-pooling. The number of output channels are 32, 64, and 128 respectively. A fully-connected layer with linear output maps the encoder output to a $M=128$ dimensional latent measurement. 
Following the structure of the encoder, we decode latent states with dimension $D=128$ by first applying a fully-connected layer with linear output, followed by 3 convolutional layers where in each layer we first upsample the input by factor 2, and then apply a $3\times 3$ filter with $1\times 1$ stride, followed by a ReLU non-linearity (for all but the last layer).

\paragraph{LGSSM}
We assume both $\mathbf{P}$ and $\mathbf{R}$ to be diagonal covariance matrices. We define $\mathbf{P}=0.1\cdot\mathbf{I}_D$, while the diagonal $\mathbf{\sigma}_R$ is learned. The latent measurement operator $\mathbf{H}$ is set to the identity matrix $\mathbf{I}_D$, assuming the latent measurement space $\mathcal{W}$ and the latent state space $\mathcal{Z}$ to have the same dimensionality $M=D=128$.
Further, we define the prior for time point $t=1$ as $\mathcal{N}(\mathbf{z}_1\mid \mathbf{0}, 10\cdot \mathbf{I}_D)$, expressing high uncertainty about the initial state.
Finally, we define the transition matrix $\mathbf{F}_t$ as
\begin{equation*}
    F_t = \sum_{k=1}^C \alpha_k(\mathbf{z}_t) \mathbf{F}^{(k)},
\end{equation*}
with $C=8$ and $\alpha_1,...,\alpha_C$ the outputs of a 2-layer MLP with softmax activation.

\paragraph{Training}
For all experiments, we trained 3 models with different random seeds for a maximum of 100 epochs using the Adam optimizer \cite{Kingma2015} with default settings. Based on a grid search, the learning rate was set to 0.001.

\subsection{Evaluation and baselines} 
We compare our method against the following baselines: A convolutional VAE \cite{Kingma2014} is used to jointly model velocities and log-densities, without capturing temporal dependencies. The encoder and decoder architecture as well as the dimensionality of the latent space matches our proposed model. Further, $\emph{velocity volume profiling}$ (VVP) \cite{Waldteufel1979} is used to reduce the set of radial velocity measurements of each radar to a single velocity vector describing a uniform velocity field around the radar. An estimate of the full velocity field is obtained by linearly interpolating the resulting $N$ velocity vectors in space. 
All methods are evaluated based on their reconstruction quality, measured by the root mean squared error (RMSE) between model reconstruction and ground-truth, for velocities and log-transformed densities respectively.

\subsection{Experiments}
\begin{figure}[htb]
    \centering
    \includegraphics[width=\textwidth]{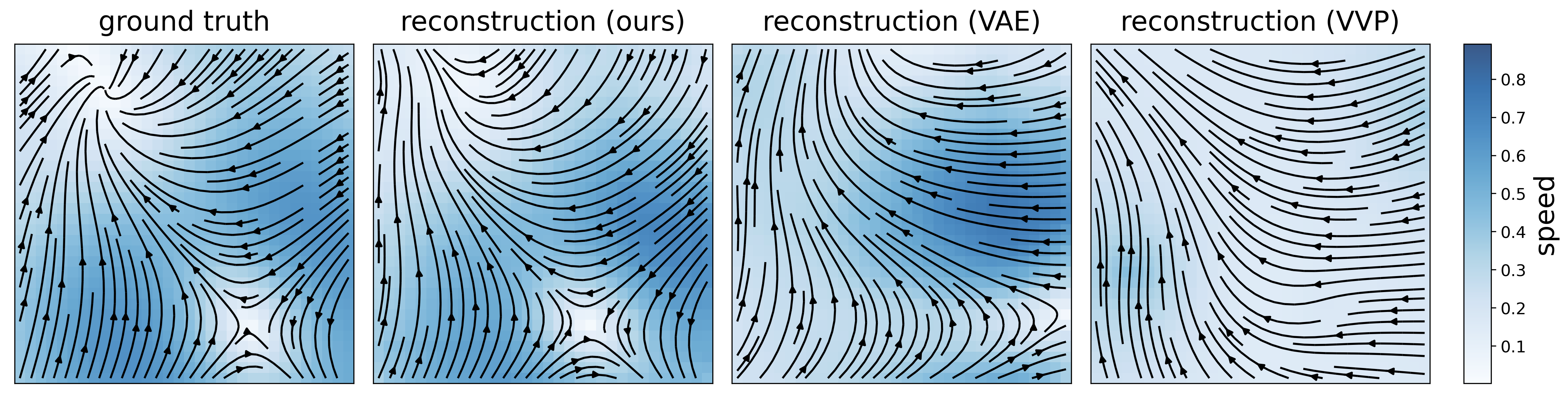}
    \caption{Qualitative comparison of three different reconstruction methods. We show reconstructions of a single velocity field (t=8) from the test set. For all methods, the radar range was set to $d_n=2$.}
    \label{fig:examples_v}
\end{figure}

\paragraph{Comparison to baselines}
We evaluate all methods for radar ranges varying from $d_n=1$ ($64 \pm 13\%$ coverage) to $d_n=2$ ($98\pm 4 \%$ coverage) to $d_n=\infty$ ($100\%$ coverage). The results, summarized in table~\ref{tab:RMSE_overview}, show that for all considered ranges our method yields the lowest RMSE for both velocities and log-densities. Fig.~\ref{fig:examples_v} visually compares the different methods based on an example velocity field from the test set. Our method is well able to capture the patterns of both direction and speed of movement, whereas the two baseline methods fail to reconstruct some of the local structures. 

\begin{table}[htb]
    \centering
    \begin{tabular}{lccccc}
        \toprule
         && $d_n$ & VVP & VAE & Ours \\
        \midrule
        \multirow{3}{*}{velocities $\mathbf{v}$}& & 1 & 0.1704 & 0.2193 \tiny{$\pm 0.0032$} & $\mathbf{0.1134}$ \tiny{$\pm 0.0048$} \\
        & & 2 & 0.1704 & 0.2139 \tiny{$\pm 0.0007$} & $\mathbf{0.0894}$ \tiny{$\pm 0.0042$} \\
        & & $\infty$ & 0.1704 & 0.2031 \tiny{$\pm 0.0046$} & $\mathbf{0.0589}$ \tiny{$\pm 0.0046$} \\
        \midrule
        \multirow{3}{*}{log-densities $\mathbf{q}$} & & 1 & -- & 0.1368 \tiny{$\pm 0.0075$} & $\mathbf{0.1144}$ \tiny{$\pm 0.0023$} \\
        & & 2 & -- & 0.1222 \tiny{$\pm 0.0031$} & $\mathbf{0.0641}$ \tiny{$\pm 0.0066$} \\
        & & $\infty$ & -- & 0.1249 \tiny{$\pm 0.0062$} & $\mathbf{0.0309}$ \tiny{$\pm 0.0018$} \\
        \bottomrule
    \end{tabular}
    \vspace{0.1cm}
    \caption{RMSE (mean $\pm$ standard deviation using 3 different seeds) for reconstructed velocities and log-densities, evaluated for three different radar ranges $d_n$.}
    \label{tab:RMSE_overview}
\end{table}

\begin{figure}[htb]
    \centering
    \includegraphics[width=0.92\textwidth]{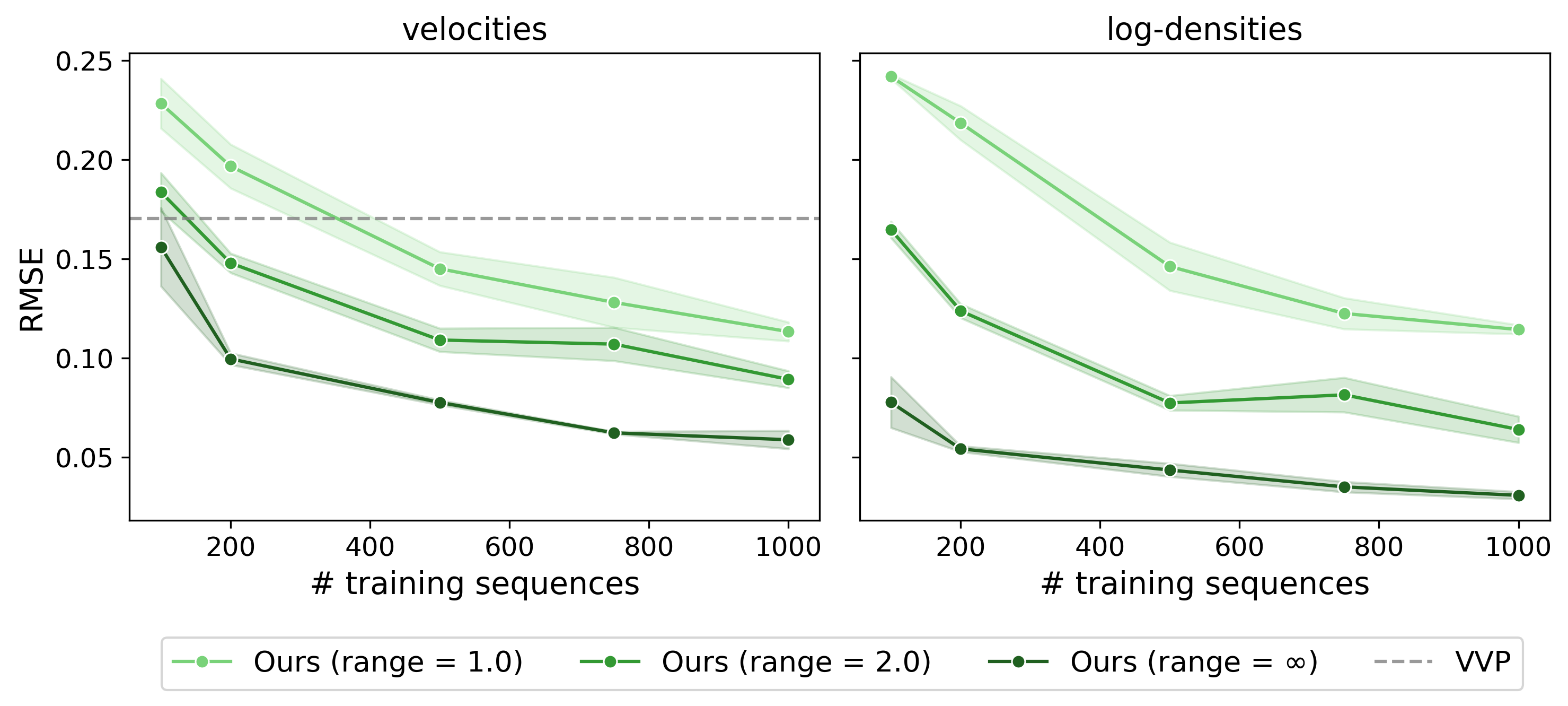}
    \caption{RMSE for velocities (left) and log-densities (right) reconstructed by our method, using a varying number of training sequences. Shaded regions correspond to one standard deviation around the mean using 3 different seeds.}
    \label{fig:vary_n_train}
\end{figure}

\paragraph{Data-efficiency}
We evaluate the data-efficiency of our method by gradually changing the number of training sequences from 100 to 1000. As before, radar ranges $d_n\in\{1,2,\infty\}$ are considered. While in general the RMSE increases as the number of training sequences decreases, we are able to obtain more accurate velocity reconstructions than the VVP baseline (which does not require any training) using as little as 200 training sequences for ranges $d_n\in\{2,\infty\}$ and 500 sequences for $d_n=1$ (see fig.~\ref{fig:vary_n_train}). In comparison, the VAE cannot beat the VVP baseline, even when trained on 1000 sequences.

\begin{figure}[htb]
    \centering
    \includegraphics[width=\textwidth]{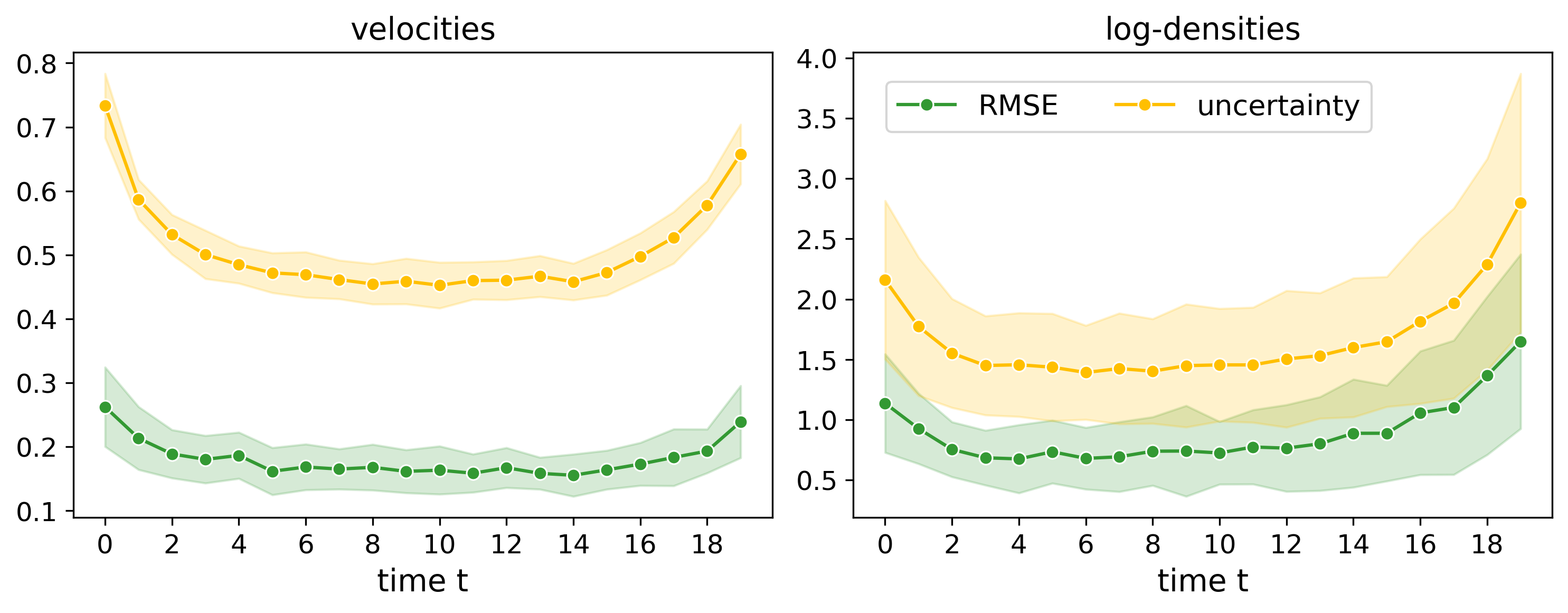}
    \caption{Reconstruction error (RMSE) and uncertainty as a function of time, using our method with radar range $d_n=2$ and 1000 training sequences. Uncertainties are computed as the standard deviation over reconstructions based on 10 samples from the latent posterior $\mathcal{N}(\mathbf{z}_t^+,\mathbf{\Sigma}_t^+)$. Shaded regions correspond to one standard deviation around the mean computed for 50 test sequences.}
    \label{fig:reconstruction_uncertainty}
\end{figure}

\paragraph{Uncertainty estimates}
Based on the estimated latent posterior $\mathcal{N}(\mathbf{z}_t^+, \mathbf{\Sigma}_t^+)$, we can quantify the uncertainty associated with the reconstructed velocity and log-density fields by repeatedly sampling $\hat{\mathbf{z}}_t\sim\mathcal{N}(\mathbf{z}_t^+, \mathbf{\Sigma}_t^+)$ and decoding $\hat{\mathbf{z}}_t$ to obtain a set of physical reconstructions. Computing the average per-grid-cell standard deviation for each time point reveals higher uncertainties at the beginning and end of each sequence, for velocities and log-densities respectively (see fig.~\ref{fig:reconstruction_uncertainty}). This matches the generally higher RMSE at the beginning and end of each sequence and indicates that our method utilizes information about both past and future to generate accurate reconstructions. 

\section{Conclusion}
In this work, we combined deep spatial feature learning with exact inference in linear Gaussian state space models to reconstruct high-dimensional time-varying animal movement patterns from partial and noisy Doppler weather radar data, without requiring in-depth knowledge about the underlying processes or access to ground-truth data. 
Our results indicate that this approach is able to reconstruct movement patterns in more detail than traditional approaches based on uniformity assumptions, while requiring less data than models neglecting spatio-temporal movement dynamics.

Nonetheless, our experiments are limited to synthetic radar data and several additional challenges need to be tackled before applying the proposed method to real weather radar data. 
Firstly, in practice biological echoes are intertwined with echoes from rain clouds, human infrastructure, etc. For our model to learn meaningful spatial features and to use mass conservation constraints effectively it is, however, crucial to reliably filter out irrelevant non-biological signals. Although there has been significant progress in in this regard, available classification methods \cite{stepanian2016dual, lin2019mistnet} remain to be adjusted and thoroughly evaluated for European weather radar data. 
Secondly, in contrast to our synthetic data generated on a 2D regular grid, real weather radar data consists of 3D polar volumes for which the resolution decreases with distance to the radar station. To apply our method, we either need to project this data onto a regular grid (and thus lose information) or adjust the encoder network to work with irregularly structured data (i.e. point clouds or graphs) directly. 
Finally, radial velocity measurements may be subject to aliasing, meaning that the true radial velocity is only known up to an additive multiple of the Nyquist velocity. Training our model based on these aliased measurements will inevitably result in erroneous reconstructions. To overcome this issue, the deterministic forward model mapping full velocity and density fields to radar measurements needs to be replaced by a probabilistic one (e.g. using a wrapped normal likelihood similar to \cite{Angell2018}).

We are hopeful that future work in these directions will eventually allow us to uncover aerial movement patterns of birds, bats and insects at various spatial scales. In the long term, this opens up opportunities for ecologists to obtain novel scientific insights into the underlying behaviours and movement strategies, and to develop more effective strategies for species conservation and mitigation of human-wildlife conflicts.

\bibliographystyle{abbrv}
\bibliography{new_references}

\newpage

\appendix

\section*{Appendix}
\section{Data generation}

We generate synthetic velocity fields by describing animal movements as the interplay between attraction to areas with food, mild climate, etc. and repulsion from geographical barriers, predators and hostile environmental conditions. That means, we assign each point in time and space a scalar value $\phi(t, \mathbf{x})$, representing the associated attractiveness (or potential). We initialize $\phi(0, \mathbf{x})$ as a mixture of 10 two-dimensional Gaussians with randomly sampled mean and diagonal covariance. Given $\phi(t, \mathbf{x})$, we generate auto-correlated potentials at time $t+1$, by shifting each mode by some fixed displacement vector.
Finally, the velocity fields associated with $\phi(t, \mathbf{x})$ are computed as $\mathbf{v}(t, \mathbf{x})=-\nabla\phi(t, \mathbf{x})$.

Similar to $\phi(0, \mathbf{x})$, we initialize animal densities as a mixture of 10 two-dimensional Gaussians. Then, we simulate eq.~\ref{eq:transport} forward in time using a \emph{forward time centered space} finite difference scheme with $dt=0.001$ to generate a time-varying density fields $\rho(t, \mathbf{x})$ that are consistent with the time-varying velocity field $\mathbf{v}$. The resulting fields are sampled at $\Delta t=0.025$ to obtain the final time series.

In practice, density distributions are highly skewed. To obtain approximately normally distributed inputs, we work with log-transformed densities instead. Before feeding $\mathbf{v}$ and $\log(\rho)$ to our model, we rescale them to values between -1 and 1. 
Measurements outside the range $d_n$ of radar $n$ are masked by setting them to zero. The corresponding vectors in the projection matrix $\mathbf{A}_n$ are set to $\mathbf{0}$ as well.

\begin{figure}[htb]
    \centering
    \includegraphics[width=1.0\textwidth]{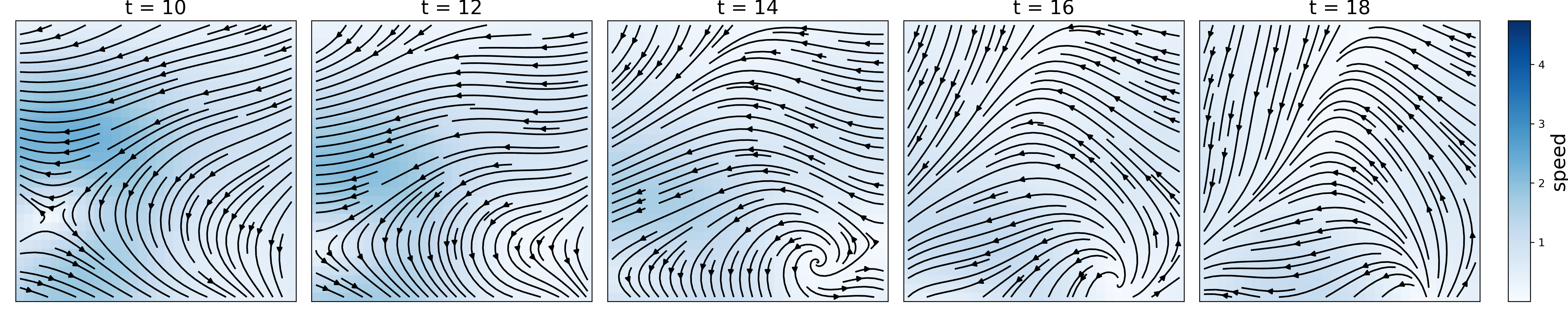}
    \includegraphics[width=1.0\textwidth]{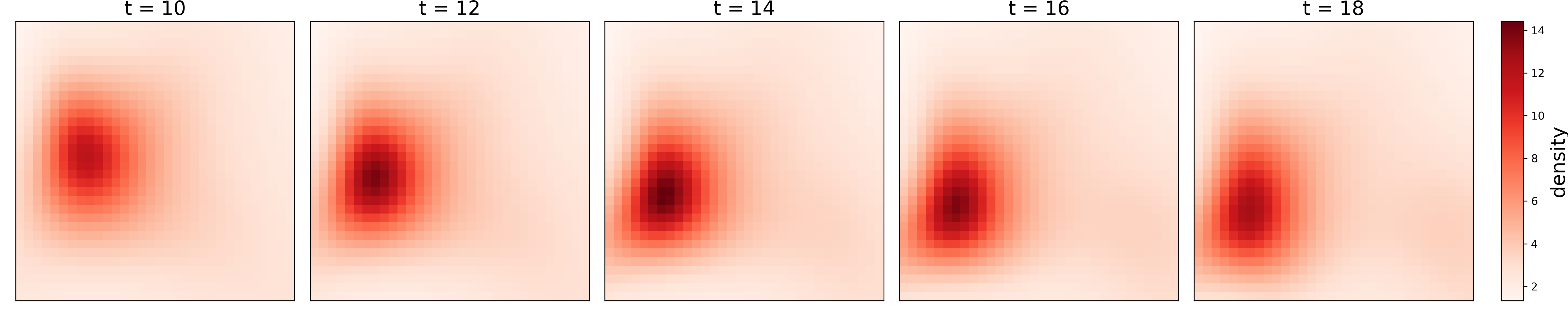}
    \caption{Samples from a time series of synthetic velocity (top row) and density (bottom row) fields.}
    \label{fig:example_ts}
\end{figure}

\begin{figure}[htb]
    \centering
    \includegraphics[width=0.9\textwidth]{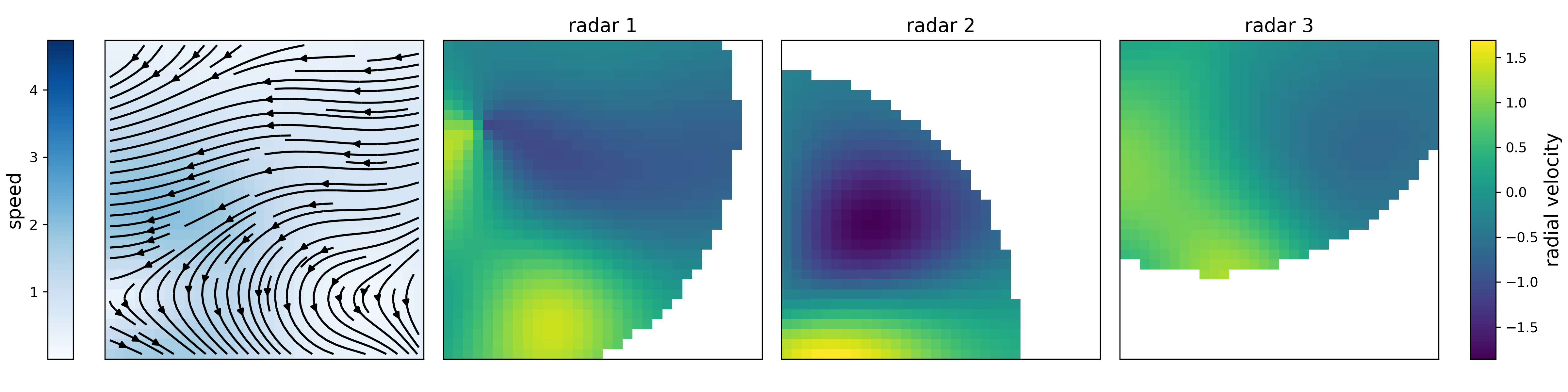}
    \caption{A full velocity field (left), and corresponding radial velocities measured by 3 different radars with range $d=2$ (right).}
    \label{fig:example_r}
\end{figure}

\section{Additional figures}

\begin{figure}[htb]
    \centering
    \includegraphics[width=1.0\textwidth]{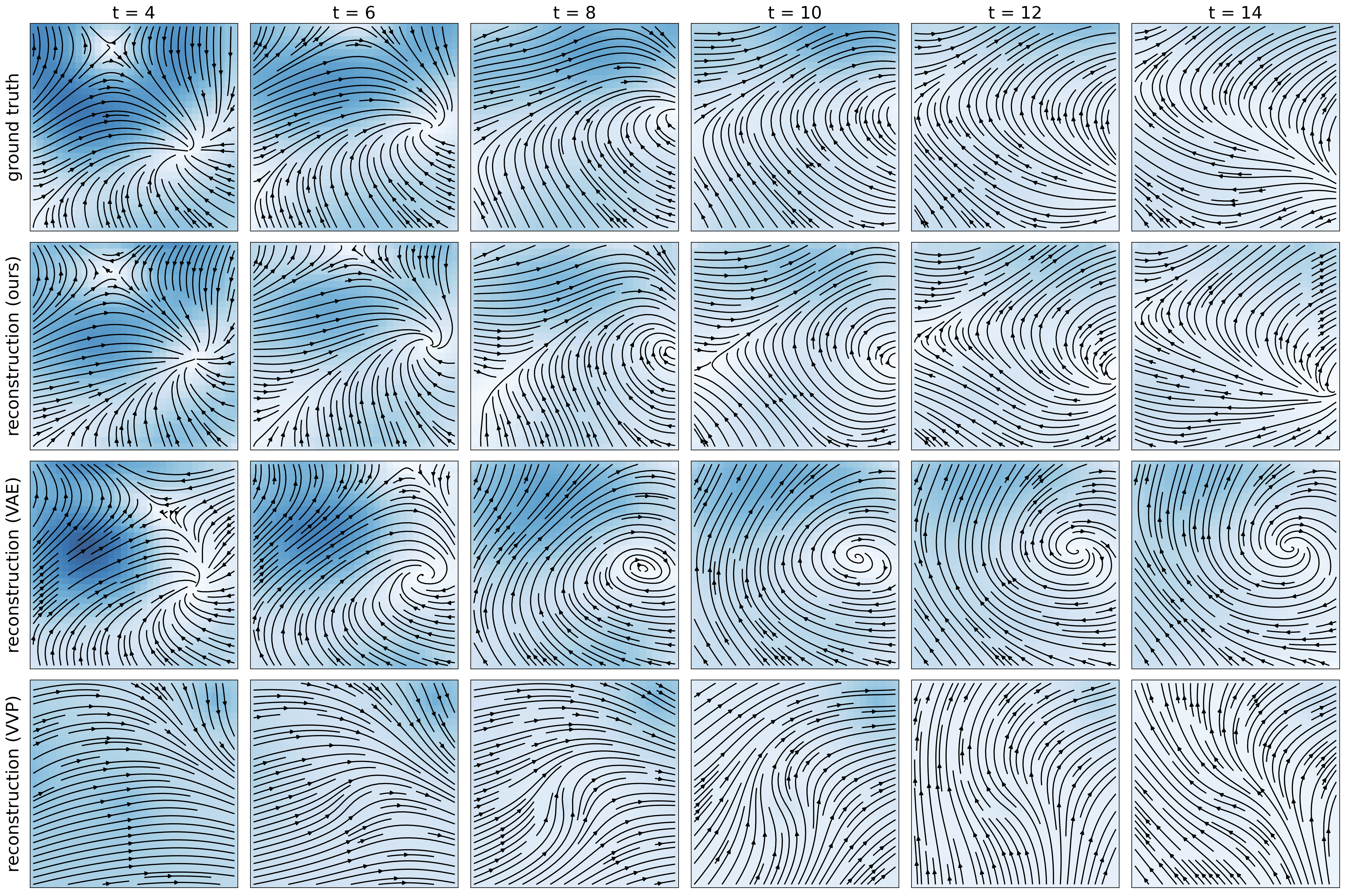}
    \includegraphics[width=1.0\textwidth]{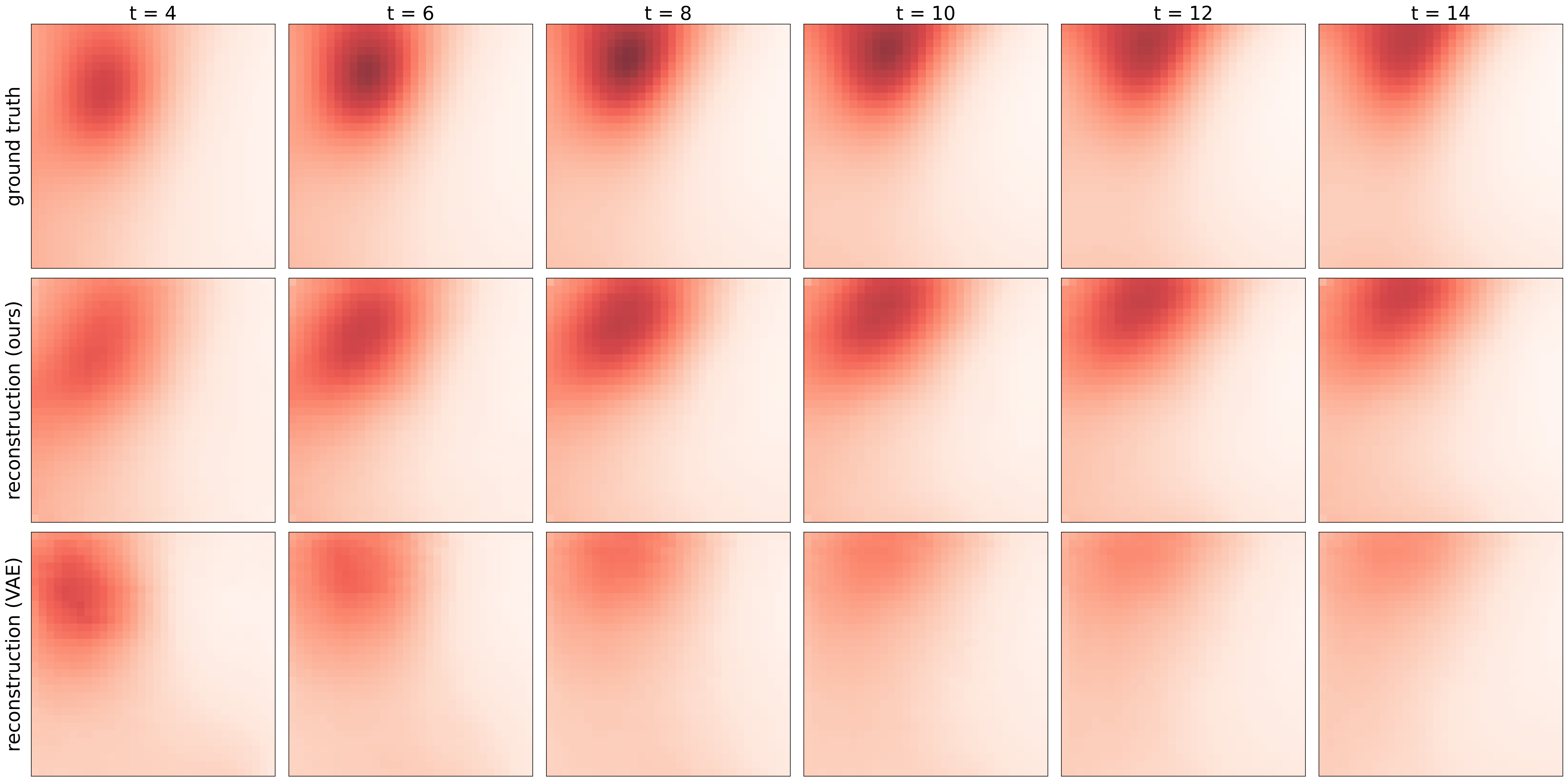}
    \caption{Example time series of synthetic velocity (blue) and density (red) fields, together with corresponding reconstructions.}
    \label{fig:example_reconstructions_ts}
\end{figure}

Fig. \ref{fig:example_reconstructions_ts} shows an example time series of ground-truth velocity and density fields together with reconstructions generated by our proposed model, a convolutional VAE, and a linear interpolation based on $\emph{velocity volume profiling}$ (VVP). Our model reconstructs both velocities and densities more accurately than the considered baseline models.

\end{document}